\documentclass[10pt,twocolumn,letterpaper]{article}

\usepackage{cvpr}
\usepackage{times}
\usepackage{epsfig}
\usepackage{graphicx}
\usepackage{amsmath}
\usepackage{amssymb}
\usepackage{bm}
\usepackage{xcolor}
\usepackage{microtype}
\usepackage{indentfirst}
\usepackage{url}
\usepackage{booktabs}
\usepackage{siunitx}

\usepackage[pagebackref=true,breaklinks=true,letterpaper=true,colorlinks,bookmarks=false]{hyperref}

\cvprfinalcopy 

\def\cvprPaperID{5723} 

\ifcvprfinal\fi
\newcommand{\idrm}{\textrm{id}}
\newcommand{\albedorm}{\textrm{albedo}}
\newcommand{\exprm}{\textrm{exp}}
\begin{document}

\title{Lightweight Photometric Stereo for Facial Details Recovery}

\author{Xueying Wang\textsuperscript{1} \quad Yudong Guo\textsuperscript{1} \quad Bailin Deng\textsuperscript{2} \quad Juyong Zhang\textsuperscript{1}\thanks{Corresponding author}\\
	\textsuperscript{1}University of Science and Technology of China \quad \textsuperscript{2}Cardiff University\\
	{\tt\small\{WXY17719, gyd2011\}@mail.ustc.edu.cn} \quad  {\tt\small DengB3@cardiff.ac.uk} \quad  {\tt\small juyong@ustc.edu.cn}
}

\maketitle

\begin{figure*}
	\centering
	\includegraphics[width=1.0\textwidth]{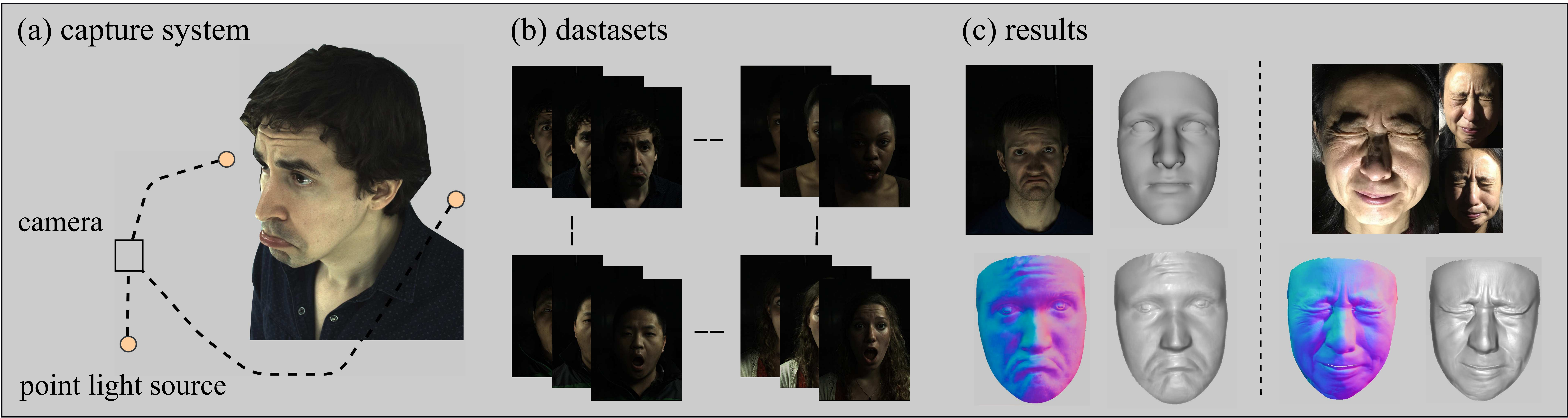}
	\caption{We propose a convolutional neural network based method for face reconstruction under photometric stereo scenario. (a) \& (b): Our dataset for network training consists of photos with different expressions, captured using a system composed of three near point light sources and a fixed camera. (c): Our proposed method can recover fine details even with a single image input (left). For images captured by a smartphone, with a hand-held light at locations not seen in our training dataset, our method also works well in this casual setup. (right).}
	\label{fig:set_up}
\end{figure*}

\begin{abstract}
Recently, 3D face reconstruction from a single image has achieved great success with the help of deep learning and shape prior knowledge, but they often fail to produce accurate geometry details. On the other hand, photometric stereo methods can recover reliable geometry details, but require dense inputs and need to solve a complex optimization problem. In this paper, we present a lightweight strategy that only requires sparse inputs or even a single image to recover high-fidelity face shapes with images captured under near-field lights. To this end, we construct a dataset containing 84 different subjects with 29 expressions under 3 different lights. Data augmentation is applied to enrich the data in terms of diversity in identity, lighting, expression, etc. With this constructed dataset, we propose a novel neural network specially designed for photometric stereo based 3D face reconstruction. Extensive experiments and comparisons demonstrate that our method can generate high-quality reconstruction results with one to three facial images captured under near-field lights. Our full framework is available at \url{https://github.com/Juyong/FacePSNet}.
\end{abstract}

\section{Introduction}

High-quality 3D face reconstruction is an important problem in computer vision and graphics~\cite{stylianou2009image} that is related to various applications such as digital actor~\cite{AlexanderRLCMWD10}, face recognition~\cite{blanz2003face,zulqarnain2018learning} and animation~\cite{AlexanderRLCMWD10,ichim2015dynamic,thies2016face2face}. Some works have been devoted to solving this problem at the source, using either multi-view information~\cite{dou2018multi,wu2019mvf} or the illumination conditions~\cite{ps3d1,ps3d2,stratou2011effect}. Although some of these methods are capable of reconstructing high-quality 3D face models with both low-frequency structures and high-frequency details like wrinkles and pores, the hardware environment is hard to set up and the underlying optimization problem is not easy to solve. For this reason, 3D face reconstruction from a single image has attracted wide attention, with many works focusing on reconstruction from an ``in-the-wild'' image~\cite{sanyal2019learning,gecer2019ganfit,deng2019accurate,Guo20193DFace}. Although most of them can reconstruct accurate low-frequency facial structures, few can recover fine facial details. In this paper, we turn our attention to the photometric stereo technique~\cite{woodham1980photometric}, and consider the near-field point light source setting due to its portability. We aim to reconstruct high-precision 3D face models with sparse inputs using photometric stereo under near point lighting.

State-of-the-art sparse photometric 3D reconstruction methods such as~\cite{cao2018sparse,chen20193d} can reconstruct 3D face shapes with fine geometric details. However, they are mainly based on conventional optimization approaches with high computational costs. In recent years, great progress has been made in deep learning-based photometric stereo~\cite{ikehata2018cnn, chen2018ps} that can estimate accurate normals. However, these existing methods cannot be directly applied to solve our problem. First, they mainly focus on general objects with dense inputs, making them not suitable for our 3D face reconstruction problem with sparse inputs. Second, they assume parallel directional lights, which is difficult to achieve in practice especially for indoor lighting conditions. To solve the sparse photometric stereo problem fast and well, we must address the following challenges. First, without the parallel lighting assumption, calibrating the lighting direction of near-field point light sources is much more complex and needs to solve a nonlinear optimization problem. Moreover, the reconstruction problem with less than three input images is ill-posed, and thus prior knowledge of the reconstruction object is needed.

In this paper, we combine deep learning-based sparse photometric stereo and facial prior information to reconstruct high-accuracy 3D face models. Currently, there is no publicly available dataset of face images captured under near point lighting conditions and their corresponding 3D geometry. Therefore, we construct such a dataset for the network training. We use real face images captured using a system composed of three near point light sources and a fixed camera. Based on this system, we develop an optimization method to recover 3D geometry along with calibrating light positions and estimating normals. Using our reconstructed 3D face models and publicly available high-quality 3D face datasets, we augment our dataset by synthesizing a large number of face images with their corresponding 3D shapes. With the real and synthetic data, we design a two-stage convolutional neural network to estimate a high-accuracy normal map from sparse input images. The coarse shape, represented by a parametric 3D face model~\cite{blanz1999morphable} and the pose parameters, are recovered in the first stage. The face images and the normal map obtained from the first stage are fed into the second-stage network to estimate a more accurate normal map. Finally, a high-quality 3D face model is recovered via a fast surface-from-normal optimization. Fig.~\ref{fig:set_up} shows the pipeline of our method. Comprehensive experiments demonstrate that our network can produce more accurate normal maps compared with state-of-the-art photometric stereo methods. Our lightweight method can also recover fine facial details better than state-of-the-art single image-based face reconstruction methods.
\section{Related work}

\textbf{Photometric Stereo.} The photometric stereo (PS) method~\cite{woodham1980photometric} estimates surface normals from a set of images captured under different lighting conditions. Since the seminal work of~\cite{woodham1980photometric}, different methods have been proposed to recover surfaces in this manner~\cite{wu2010robust,herbort2011introduction}. Many such methods assume directional lights with infinite light source positions. On the other hand, some works focus on reconstruction under near point light sources, using optimization approaches that are often complex and time-consuming~\cite{xie2015photometric,liu2018near,queau2018led}. 
To achieve efficiency for practical applications with near point light sources, we only adopt optimization-based methods to construct the training dataset and then train the neutral model for lightweight photometric stereo for 3D face reconstruction. The most related work to our training data construction step is~\cite{cao2018sparse}, which proposed an iteration pipeline to reconstruct high-quality 3D face models.

\textbf{Deep Learning-Based Photometric Stereo.} With the development of convolutional neural networks, various deep learning-based approaches have been proposed to solve photometric stereo problems. Most of them can be categorized into two types according to their input. The first type requires images together with corresponding calibrated lighting conditions. Santo \etal~\cite{santo2017deep} proposed a differentiable multi-layer deep photometric stereo network (DPSN) to learn the mapping from a measurement of a pixel to the corresponding surface normal. Chen \etal~\cite{chen2018ps} put forward a fully connected convolutional network to predict the normal map of a static object from an arbitrary number of images. A physics-based unsupervised neural network was proposed by Taniai \etal~\cite{taniai2018neural} with both surface normal map and synthesized images as output. Ikehata~\cite{ikehata2018cnn} presented an observation map to describe pixel-wise illumination information, and estimated surface normals with the observation map as input to an end-to-end convolutional network. Furthermore, Zheng \etal~\cite{zheng2019spline} and Li \etal~\cite{li2019learning} solved the sparse photometric stereo problem based on the observation map. This type of work assumes lighting directions as prior and cannot handle unknown lighting directions. 
The second type directly estimates lighting conditions and normal maps altogether from the input images. A network named UPS-FCN was introduced in~\cite{chen2018ps} to calibrate lights and predict surface normals. Later, Chen \etal~\cite{chen2019self} proposed a two-stage deep learning architecture called SDPS-Net to handle this uncalibrated problem. Both types focus on solving photometric stereo problems under directional lights which is difficult to achieve in practice, and most of these methods do not perform well with sparse inputs. In this paper, we solve the sparse uncalibrated photometric stereo problem under near-field point light sources.

\textbf{Single Image-Based 3D Face Reconstruction.} 3D face reconstruction from a single image has made great progress in recent years. The key to this task is to establish a correspondence map from 2D pixels to 3D points. Jackson \etal~\cite{jackson2017large} proposed to directly regress a volumetric representation of the 3D mesh from a single face image with a convolutional neural network. Feng \etal~\cite{feng2018joint} designed a 2D representation called UV position map to record 3D positions of a complete human face. Deng \etal~\cite{deng2019accurate} directly regressed a group of parameters based on 3DMM~\cite{blanz1999morphable,booth20163d,paysan20093d}. All these works can reconstruct the 3D face model from a single image but cannot recover geometry details. Recently, this issue has been addressed with a \emph{coarse-to-fine} reconstruction strategy. Sela \etal~\cite{sela2017unrestricted} first constructed a coarse model based on a depth map and a dense correspondence map and then recovered details in a geometric refinement process. Richardson \etal~\cite{richardson2017learning} developed an end-to-end CNN framework composed of a CoarseNet and a FineNet to reconstruct detailed face models. Jiang \etal~\cite{jiang20183d} designed a three-stage approach based on a bilinear face model and the shape-from-shading (SfS) method. Li \etal~\cite{li2018feature} recovered face details using SfS along with an albedo prior mask and a depth-image gradient constraint. Tran \etal~\cite{tran2018extreme} proposed a bump map to describe face details and use a hole filling approach to handle occlusions. Chen \etal~\cite{chen2019photo} recovered high-quality face models based on a proxy estimation and a displacement map. For 3D face reconstruction from caricature images, Wu \etal~\cite{WuZLZC18} proposed an intrinsic deformation representation for extrapolation from normal 3D face shapes.

Most existing works approximated the human face as a Lambertian surface and simulated the environment light using the spherical harmonics (SH) basis functions, which is not suitable for the near point lighting condition due to a large area of shadows. Based on our constructed dataset, we also design a network that can reconstruct a 3D face model with rich details from a single image captured under the near point lighting condition.
\section{Dataset Construction}
\label{sec:dataset}

In this paper, we propose a lightweight method to reconstruct high-quality 3D face models from uncalibrated sparse photometric stereo images. As there is no publicly available dataset that contains face images with near point lighting and their corresponding 3D face shapes, we construct such a dataset by ourselves. Given face images captured under different light sources, we would like to solve for the albedos and the normals of the face model such that the intensities of the resulting images under calibrated lights are consistent with the observed intensities from the input images. This problem may be ill-posed with only three input images due to the presence of shadows. Therefore, we utilize a parametric 3D face model as prior knowledge, and propose an optimization method to estimate accurate normal maps. In this section, we first introduce some related basic knowledge, and then present how we construct the real image-based dataset and synthetic dataset.

\subsection{Preliminaries}
\label{sec:preliminary}

{\bf Imaging Formula}. We approximate the human face as a Lambertian surface and simulate the near point lighting condition using the photometric stereo. Given a point light source at position $\mathbf{P}_j \in \mathbb{R}^{3}$ with illumination $\beta_j \in \mathbb{R}$, the imaging formula for a point $i$ can be expressed as~\cite{xie2015photometric}:
\begin{equation}
    \label{eq:imaging_formula_1}
    \mathbf{I}_{ij}(\mathbf{V}_i, \mathbf{N}_i, \boldsymbol{\rho}_i) \triangleq  \boldsymbol{\rho}_i \left(\mathbf{N}_i \cdot \frac{\beta_j\left(\mathbf{P}_j - \mathbf{V}_i\right)}{\left\| \mathbf{P}_j - \mathbf{V}_i \right\|_{2}^{3}}\right),
\end{equation}
where $\mathbf{V}_i, \mathbf{N}_i \in \mathbb{R}^{3}$ are the position and normal of the point, and $\mathbf{I}_{ij}, \boldsymbol{\rho}_i \in \mathbb{R}^{3}$ are the intensity and albedo in the RGB color space, respectively. Given the captured images, the photometric stereo problem with near point light sources is to recover lighting positions and illuminations, the vertex position, albedo and normal of a point on the object.

{\bf Parametric Face Model}. 3DMM~\cite{blanz1999morphable} is a widely used parametric model for human face geometry and albedo. We use 3DMM to build a coarse face model for further optimization. In general the parametric model represents the face geometry $\mathbf{G} \in \mathbb{R}^{3n_v}$ and albedo $\mathbf{A} \in \mathbb{R}^{3n_v}$ as
\begin{align}
    \mathbf{G} &= \overline{\mathbf{G}} + \mathbf{B}_{\idrm}\boldsymbol{\alpha}_{\idrm} + \mathbf{B}_{
    	\exprm}\boldsymbol{\alpha}_{\exprm}, \label{eq:geo_3dmm}\\
    \mathbf{A} &= \overline{\mathbf{A}} + \mathbf{B}_{\albedorm}\boldsymbol{\alpha}_{\albedorm},
\end{align}
where $n_v$ is the number of vertices of the face model; $\overline{\mathbf{G}} \in \mathbb{R}^{3n_v}$ and $\overline{\mathbf{A}} \in \mathbb{R}^{3n_v}$ are respectively the mean shape and albedo; $\boldsymbol{\alpha}_{\idrm} \in \mathbb{R}^{100}$, $\boldsymbol{\alpha}_{\exprm} \in \mathbb{R}^{79}$ and $\boldsymbol{\alpha}_{\albedorm}\in \mathbb{R}^{100}$ are corresponding coefficient parameters specifying an individual; $\mathbf{B}_{\idrm} \in \mathbb{R}^{3n_v\times100}$, $\mathbf{B}_{\exprm} \in \mathbb{R}^{3n_v\times79}$ and $\mathbf{B}_{\albedorm} \in \mathbb{R}^{3n_v\times100}$ are principle axes extracted from some 3D face models by PCA. We use the Basel Face Model (BFM)~\cite{paysan20093d} for $\mathbf{B}_{\idrm}$ and $\mathbf{B}_{\albedorm}$, and the FaceWarehouse~\cite{cao2013facewarehouse} for $\mathbf{B}_{\exprm}$.

{\bf Camera Model}. We use the standard perspective projection to project the 3D face model to the image plane, which can be expressed as
\begin{equation}
    \mathbf{q}_i = \boldsymbol{\Pi}(\mathbf{R}\mathbf{V}_i + \mathbf{t}),
\end{equation}
where $\mathbf{q}_i \in \mathbb{R}^2$ is the location of vertex $\mathbf{V}_i$ in the image plane, and $\mathbf{R} \in \mathbb{R}^{3\times3}$ is the rotation matrix constructed from Euler angles \emph{pitch}, \emph{yaw} and \emph{roll}, $\mathbf{t} \in \mathbb{R}^{3}$ is the translation vector, and $\boldsymbol{\Pi}: \mathbb{R}^3 \rightarrow \mathbb{R}^2$ is the perspective projection.

\subsection{Construction of Real Dataset}
\label{sec:optimization}
\begin{figure}[t!]
	\centering
	\includegraphics[width=1.0\columnwidth]{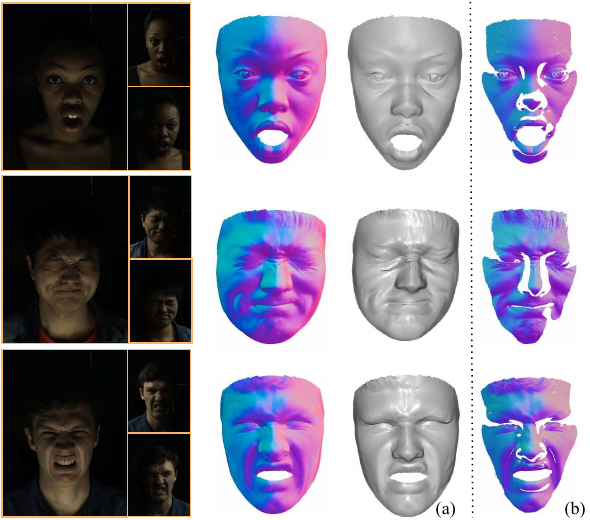}
	\caption{(a) Some results of our constructed real dataset. From left to right: input images, estimated normals and the reconstructed face models. (b) Updated normals with method in~\cite{cao2018sparse} that chooses at least three reliable lights to update normals after handling shadows. This method can only update a part of normals due to the large area of shadows and only three input images. Thus it is not suitable for our situation.}
	\label{fig:real_dataset}
\end{figure}

\begin{figure}[t!]
	\centering
	\includegraphics[width=0.5\textwidth]{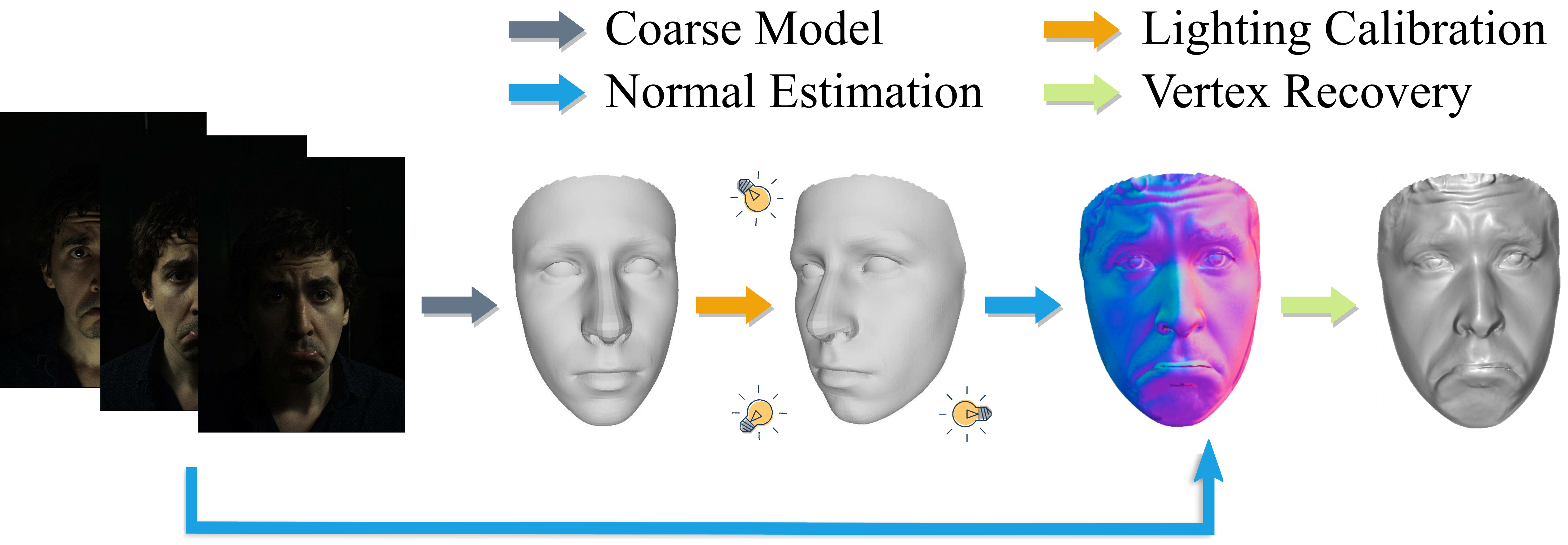}
	\caption{The algorithm pipeline of real dataset construction.}
	\label{fig:optimization_pipeline}
\end{figure}

Our real dataset is derived from photometric face images captured using a system consisting of three near point light sources (on the front, left and right) and a fixed camera. The dataset contains 84 subjects covering different races, genders and ages, with each subject captured under 29 different expressions. All images are captured at the resolution of $1600 \times 1200$. Similar to~\cite{cao2018sparse}, we design an optimization-based method to reconstruct a 3D face model with rich details from a set of images captured under different near point lighting positions and illuminations. The method in~\cite{cao2018sparse} uses the face shape prior for lighting calibration, then estimates the normals and recovers the depths in the image plane. Different from existing photometric stereo methods which always need more than three images, we have only three images as input and there may exist under-determined parts caused by shadows (Fig.~\ref{fig:real_dataset}~(b)). To alleviate this problem, we utilize the parametric model to help recover the normals. From the recovered coarse shape and updated normals, we can recover the 3D face shape with fine details as shown in Fig.~\ref{fig:real_dataset}~(a). Our algorithm pipeline is shown in Fig.~\ref{fig:optimization_pipeline}.

In order to provide a good initial 3D face shape for the following optimization, we first generate the coarse face model with three image inputs using the optimization-based inverse rendering method in~\cite{jiang20183d}. Different from the problem setting in~\cite{jiang20183d} which has only one input image, we have three face images that share the same shape, expression and albedo parameters but with different lighting conditions. After recovering the coarse face model, we calibrate the light positions $\mathbf{P} \in \mathbb{R}^{3 \times n}$ and illuminations $\boldsymbol{\beta} \in \mathbb{R}^n$ using the calibration method proposed in~\cite{cao2018sparse}. Since the Lambertian surface model is invalid in regions under shadows, we use a simple filter to determine the available light sources $\mathcal{L}_i$ for each triangle of the 3D face mesh by
\begin{equation}
    \label{eq:ava_lights}
    \mathcal{L}_i = \left\{ j \mid \mathbf{N}_i^f \cdot (\mathbf{P}_j - \mathbf{V}_i^f) > 0, j = 1, \ldots, n\right\}
\end{equation}
where $\mathbf{P}_j \in \mathbb{R}^{3}$ is the position of the $j^{\textrm{th}}$ light source, and $\mathbf{N}_i^f, \mathbf{V}_i^f \in \mathbb{R}^{3}$ are the normal and centroid of the $i^{\textrm{th}}$ triangle. We only use available light sources in $\mathcal{L}_i$ for each triangle to update its normal. During photometric stereo optimization, we first optmize the triangle normals and then recover the vertex positions from the updated normals.

\paragraph{Normal Update.} As the 3D face mesh recovered by the parametric model only contains low-frequency signals, rich geometry details are lost. Thus we refine the normal of each triangle based on the photometric stereo. The updated normal $\hat{\mathbf{N}}$ and albedo $\hat{\boldsymbol{\rho}}$ are optimized via:
\begin{align}
    \min\limits_{\hat{\boldsymbol{\rho}}, \hat{\mathbf{N}}}~  
    & \sum_{i\in {F}_v}\sum_{j \in \mathcal{L}_i} \left\|\overline{\mathbf{I}}_{ij} - {\mathbf{I}}_{ij}(\mathbf{V}_i^f, \hat{\mathbf{N}}_i^f, \hat{\boldsymbol{\rho}}_i^f)\right\|_2^2  \nonumber\\
    &  + \mu_1\left\|\hat{\mathbf{N}} - {\mathbf{N}}\right\|_F^2 + \mu_2\sum\limits_{i \in \mathcal{F}_v}\left\| \hat{\boldsymbol{\rho}}_i^f - \frac{1}{|\Omega_i|}\sum_{j \in \Omega_i}\hat{\boldsymbol{\rho}}_j^f \right\|_2^2 \nonumber\\
    \textrm{s.t.}~
    & \left\| \hat{\mathbf{N}}_i^f \right\|_2 = 1 \ (i = 1, \ldots, |\mathcal{F}_v|).
    \label{eq:normal}
\end{align}
Here the first term penalizes the deviation between the observed intensity $\overline{\mathbf{I}}_{ij}$ from the input images and the intensity resulting ${\mathbf{I}}_{ij}$ evaluated with Eq.~\eqref{eq:imaging_formula_1} using the updated albedo $\hat{\boldsymbol{\rho}}_i^f$  and the updated normal $\hat{\mathbf{N}}_i^f$  at each triangle centroid $\mathbf{V}_i^f$, with $\mathcal{F}_v$ representing the set of visible triangles on the initial face model. $\overline{\mathbf{I}}_{ij}$ is determined by projecting the centroid $\mathbf{V}_i^f$ onto the image plane and performing bilinear interpolation of its nearest pixels.
The second term penalizes the deviation between the updated normals $\hat{\mathbf{N}} \in \mathbb{R}^{3 \times |\mathcal{F}_v|}$ on visible triangles and the corresponding normals ${\mathbf{N}} \in \mathbb{R}^{3 \times |\mathcal{F}_v|}$ on the initial face model.
The last term regularizes the smoothness of the updated albedo, with $\Omega_i$ denoting the set of visible triangles in the one-ring neighborhood of triangle $i$. 
We solve Eq.~\eqref{eq:normal} via alternating minimization. Specifically, we optimize $\hat{\mathbf{N}}$ while fixing $\hat{\boldsymbol{\rho}}$, and then optimize $\hat{\boldsymbol{\rho}}$ while fixing $\hat{\mathbf{N}}$. This process is iterated until convergence.

\paragraph{Vertex Recovery.}
After updating the triangle normals $\hat{\mathbf{N}}$, we optimize the face shape as a height filed $\mathbf{Z} \in \mathbb{R}^{m}$ over the image plane to match the updated normals, where $m$ is the number of the pixels covered by the projection of the coarse face model. We first transfer $\hat{\mathbf{N}}$ to pixel normals via the standard perspective projection. Then we compute $\mathbf{Z}$ via:
\begin{equation}
\min_{\mathbf{Z}}~~\left\|\widetilde{\mathbf{N}} - {\mathbf{N}}^0 \right\|_F^2 + w_1\left\| \mathbf{Z} - \mathbf{Z}^0 \right\|_2^2 + w_2\left\| \Delta \mathbf{Z} \right\|_2^2.
\label{eq:VertexRecoveryOpt}
\end{equation}
Here $\mathbf{Z}^0 \in \mathbb{R}^m$ is the initial height field obtained from the coarse face model.
$\Delta \mathbf{Z} \in \mathbb{R}^{m}$ denotes the Laplacian of the height field, and the third term in Eq.~\eqref{eq:VertexRecoveryOpt} is to regularize the smoothness of height field.
${\mathbf{N}}^0,  \widetilde{\mathbf{N}} \in \mathbb{R}^{3 \times m}$ collect the pixel normals derived from the triangle normals $\hat{\mathbf{N}}$ and from the height field $\mathbf{Z}$, respectively. Specifically, to derive the normal $\mathbf{N}_p$ for a pixel $p$ from the height field, we first project the pixel back into its 3D location $\mathbf{V}_p$ by inverting the standard perspective projection. Then $\mathbf{N}_p$ is computed as
\begin{equation*}
\begin{aligned}
\mathbf{N}_{p} = \frac{\mathbf{e}_2\times\mathbf{e}_1 + \mathbf{e}_3\times\mathbf{e}_2 + \mathbf{e}_4\times\mathbf{e}_3 + \mathbf{e}_1\times\mathbf{e}_4}{\|\mathbf{e}_2\times\mathbf{e}_1 + \mathbf{e}_3\times\mathbf{e}_2 + \mathbf{e}_4\times\mathbf{e}_3 + \mathbf{e}_1\times\mathbf{e}_4\|},
\end{aligned}
\end{equation*}
where $\mathbf{e}_1, \mathbf{e}_2, \mathbf{e}_3, \mathbf{e}_4$ denote the vectors from $\mathbf{V}_p$ to the 3D locations of $p$'s four neighbor pixels in counter-clockwise order. This non-linear least squares problem is solved with Gauss-Newton algorithm.

\begin{figure}[t!]
	\centering
	\includegraphics[width=1.0\columnwidth]{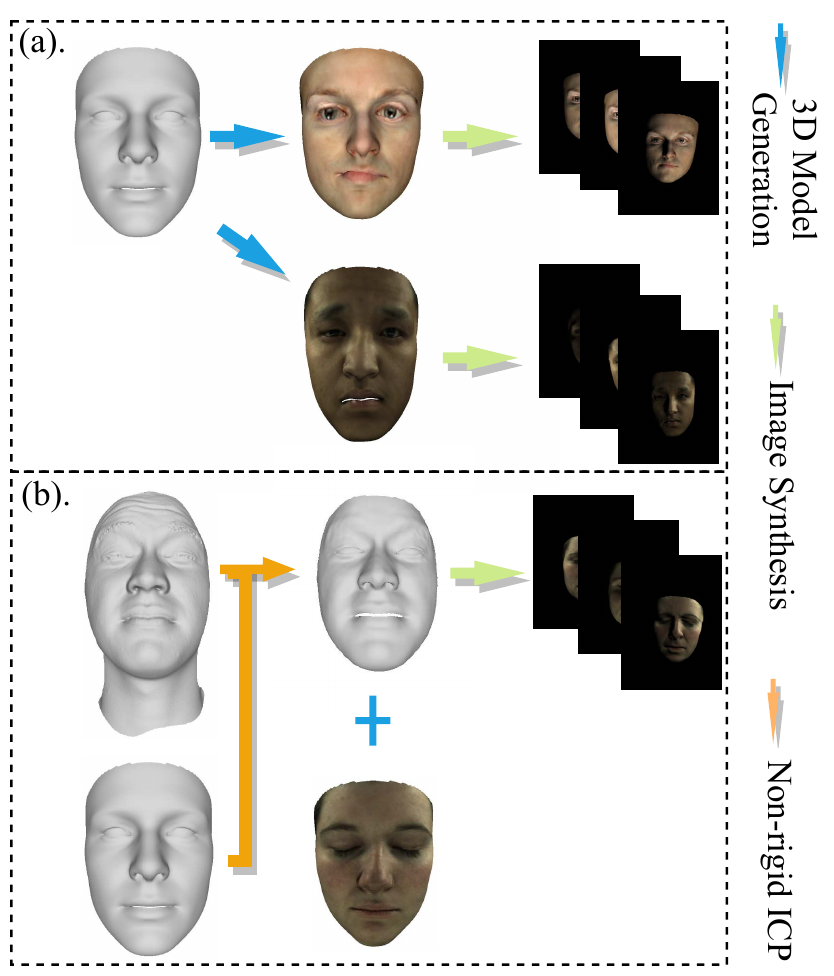}
	\caption{The process of our data augmentation methods. (a) We generate different geometries by randomly generating shape and expression parameters from 3DMM~\cite{blanz1999morphable} and transfer albedos obtained in our real dataset. (b) We use non-rigid ICP~\cite{amberg2007optimal} to fit the face models in Light Stage~\cite{ma2007rapid} with the mean shape, together with albedos in our real dataset to generate training data.}
	\label{fig:data_enhancement}
\end{figure}


\subsection{Construction of Synthetic Dataset}
\label{sec:data_aug}

To improve the coverage of our dataset, we further construct a synthetic dataset. We use albedos and 3D face models obtained from the Light Stage~\cite{ma2007rapid}, a publicly available dataset containing 23 people with 15 different expressions and their corresponding high-resolution 3D models, as the ground truth. Then we render synthetic images under three random point light positions and illuminations calibrated from our real dataset using Eq.~\eqref{eq:imaging_formula_1}. 

\noindent
\textbf{Data augmentation.} In order to fit the requirement of further network training we carry out a data augmentation process mainly from the following two aspects. On the one hand, we use the parametric model introduced in Sec.~\ref{sec:preliminary} to present different face geometry structures and albedos by randomly generating parameters  \{$\boldsymbol{\alpha}_{\idrm}, \boldsymbol{\alpha}_{\exprm}, \boldsymbol{\alpha}_{\albedorm}$\}. We transfer the albedos obtained from our real dataset to such shape models with randomly generated shape parameters, since our initial coarse model is based on the same topology. On the other hand, to have accurate parametric models as ground truth for network training on our synthetic dataset, we register a neutral parametric model to 3D face models obtained from the Light Stage using the non-rigid ICP~\cite{amberg2007optimal}, and find closest points between these two types of models as their correspondence. We further transfer albedos in our real dataset according to this correspondence. After generating those mentioned models,
 we render three images for each model with point light sources calibrated in our real dataset. The process is shown in Fig.~\ref{fig:data_enhancement}.
\section{Deep Photometric Stereo for 3D Faces}

\begin{figure*}
    \centering
    \includegraphics[width=1.0\textwidth]{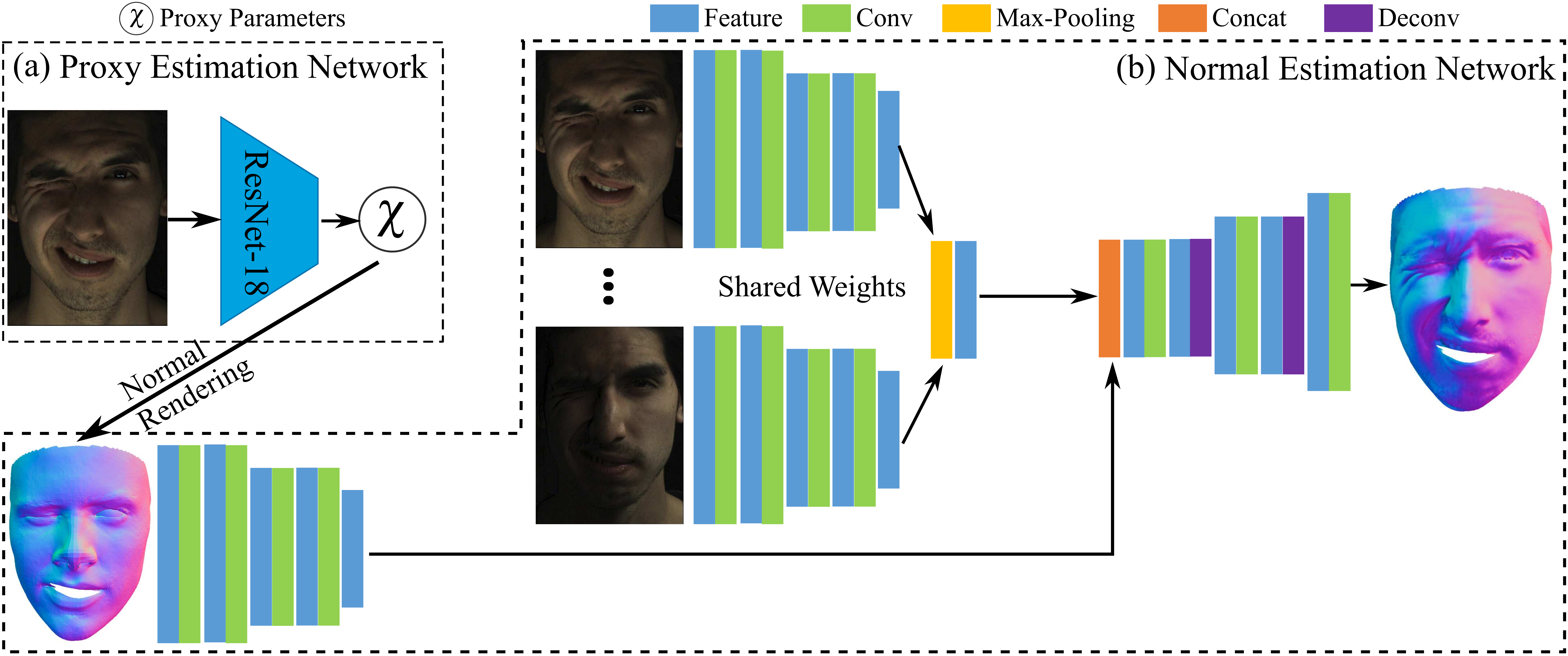}
    \caption{The architecture of our two-stage network which consists of (a) Proxy Estimation Network and (b) Normal Estimation Network. The connection between the two modules is a rendering layer which generates a coarse normal map with the estimated proxy parameters.}
\label{fig:network}
\end{figure*}

The optimization-based method described in Sec.~\ref{sec:optimization} can recover high-quality facial geometry from several face images captured under different point lighting conditions, but the procedure is time-consuming and requires at least three images as input due to the ambiguity of geometry and albedo. To alleviate these problems, we propose a CNN-based method to learn high-quality facial details from an arbitrary number of face images captured under different near point lighting conditions. Similar to the procedure in Sec.~\ref{sec:optimization}, we use a two-stage network to regress a coarse face model represented with 3DMM and a high-quality normal map respectively. With the power of CNN and our well-constructed dataset, our method can efficiently recover high-quality facial geometry even with a single image, which is not possible for optimization-based photometric stereo methods and other deep photometric stereo methods that do not utilize facial priors. Better results can be obtained with more input images. The network structure is shown in Fig.~\ref{fig:network}.

\subsection{Proxy Estimation Network}

At the first stage, we learn the 3DMM parameters and pose parameters directly from a single image to obtain a coarse face model as a proxy for the second stage with a ResNet-18~\cite{he2016deep}. The set of regressed parameters is represented by $\bm{\chi} = \{\bm{\alpha}_{\idrm}, \bm{\alpha}_{\exprm}, pitch, yaw, roll, \mathbf{t}\}$. To train the proxy estimation network, we use both the real data and the synthetic data with ground truth parameters as described in Sec.~\ref{sec:dataset}. To enrich the data, we also synthesize $5000$ images using the data augmentation strategy described in Sec.~\ref{sec:data_aug}. 

We use two loss terms to evaluate the alignment of dense facial geometry and sparse facial features respectively. The first term computes the distance between the recovered geometry and the ground truth geometry as follows:
\begin{equation}
E_{\textrm{geo}}(\bm{\chi}) = \|\mathbf{G} - \mathbf{G}_{\textrm{gt}}\|_2^2,
\end{equation}
where $\mathbf{G}$ is the geometry recovered with Eq.~\eqref{eq:geo_3dmm} and $\mathbf{G}_{\textrm{gt}}$ is the ground truth geometry. As facial landmarks convey the structural information of the human face, we design the second term to measure how close the projected 3D landmark vertices are to the corresponding landmarks in the imge:
\begin{equation}
E_{\textrm{lan}}(\bm{\chi}) = \frac{1}{|\mathcal{L}|}\sum_{i \in \mathcal{L}}\|\mathbf{q}_{i} - \mathbf{\Pi}(\mathbf{R}\mathbf{V}_{i} + \mathbf{t})\|_{2}^{2},
\end{equation}
where $\mathcal{L}$ is the set of landmarks, $\mathbf{q}_{i}$ is a detected landmark position in the input image, and $\mathbf{V}_{i}$ is the corresponding vertex location in the 3D mesh. The final loss function is a combination of the two loss terms:
\begin{equation}
E_{\textrm{loss}}(\bm{\chi}) = E_{\textrm{geo}}(\bm{\chi}) + w_{\textrm{lan}}E_{\textrm{lan}}(\bm{\chi})
\end{equation}
where $w_{\textrm{lan}}$ is a tuning weight.

\subsection{Normal Estimation Network}

The recovered geometry at the first stage lacks facial details due to the limited representation ability of 3DMM. To recover the facial geometry with final details, we learn an accurate normal map by utilizing the appearance information from face images and the geometric information from the proxy model obtained at the first stage. Specifically, the input to our normal estimation network is several face images and the normal map rendered with parameters obtained from our proxy estimation network, and the output is a refined normal map that contains high-quality facial details. The network architecture is similar to PS-FCN~\cite{chen2018ps}, which consists of a shared-weight feature extractor, an aggregation layer, and a normal regression module. One notable difference is that PS-FCN requires lighting information as input, while our normal estimation network requires proxy geometry as input to utilize facial priors. The loss function for normal estimation network is:
\begin{equation}
E_{\textrm{normal}} = \frac{1}{|\mathcal{M}|}\sum_{i \in \mathcal{M}}(1-\mathbf{n}_i^T\hat{\mathbf{n}}_{i}), 
\end{equation}
where $\mathcal{M}$ is the set of all pixels in the face region covered by the coarse face model, $\mathbf{n}_i$ and $\hat{\mathbf{n}}_{i}$ is the estimated and ground truth normals at pixel $i$, respectively.

With the estimated accurate normal map, we then obtain a high-quality face model using the vertex recovery method as explained in Sec.~\ref{sec:optimization}.
\section{Experiments}

\begin{figure*}
	\centering
	\includegraphics[width=1.0\textwidth]{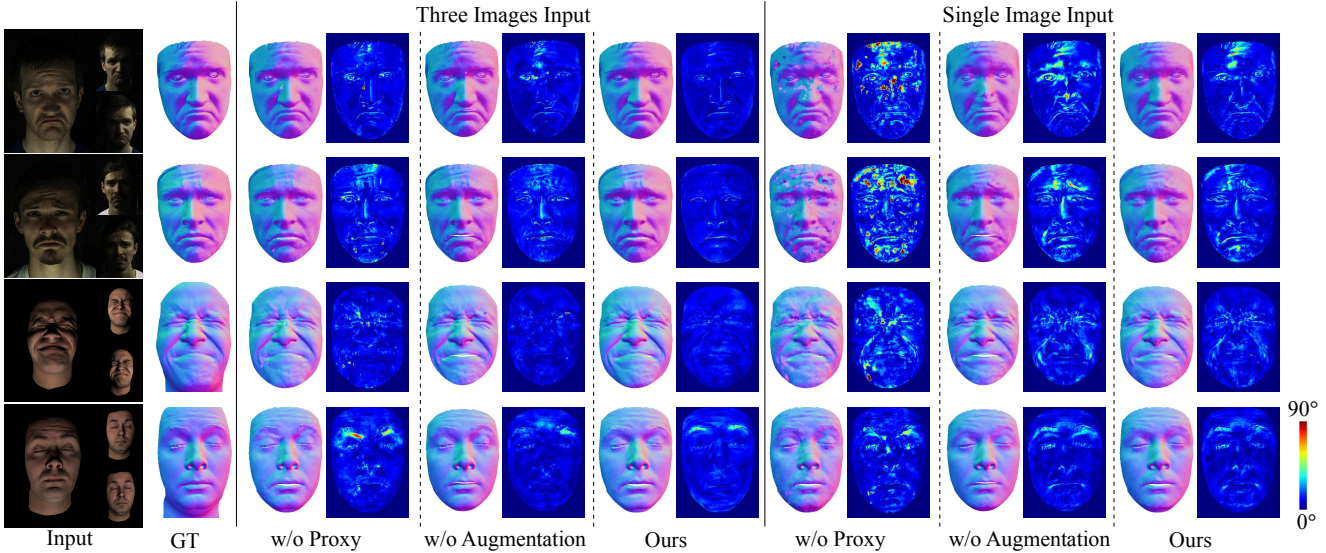}
	\caption{Ablation studies that compare the proposed method with two approaches that exclude the proxy estimation module and data augmentation respectively. For each method, we show the estimated normal maps and the corresponding angular error maps. And we use the leftmost images for single image input.}
	\label{fig:ablation_study}
\end{figure*}

\subsection{Implementation Details}
\label{sec:Imp_details}
To evaluate the proposed method, we select $77$ subjects from our captured dataset and $18$ subjects from the Light Stage dataset to train our networks and use the other subjects for testing, yielding $95$ subjects with $2503$ samples for training and $12$ subjects ($7$ from our constructed dataset and $5$ from the Light Stage dataset) with $278$ samples for testing.
We implement our method in PyTorch~\cite{paszke2017automatic} and optimize the networks' parameters with Adam solver~\cite{kingma:adam}. We first train the proxy estimation network for $200$ epochs with a batch size of $50$. Then we train the normal estimation network for $100$ epochs with a batch size of $6$ for an arbitrary number of input images. Specifically, we randomly choose one, two or three images as input in every mini-batch during training. It takes about one hour to train the proxy estimation network and $12$ hours to train a normal estimation network on a single RTX 2080 Ti GPU. The results on our test set with different inputs are shown in Tab.~\ref{table:nettest}. It can be seen that better results are achieved with more input images.

\begin{table}[!t]
	\scriptsize
	\centering  
	\caption{Average angular errors (in degrees) on test set with different inputs. S1, S2, S3 represent the leftmost, the upper-right corner and the lower-right corner image respectively.}	
	\label{table:nettest}
	\centering
	\begin{tabular}{ccccccc}  
		\toprule   
		S1 & S2 & S3 & S1\&S2 & S2\&S3 & S3\&S1 & S1\&S2\&S3 \\    
		\midrule   
		10.641 & 10.635 & 10.705 & 8.245 & 8.476 & 8.328 & 6.498 \\  
		\bottomrule  
	\end{tabular}
\end{table}

\subsection{Ablation Study}

To validate the design of our architecture, we compare the proposed method with alternative strategies that exclude some components. First, we demonstrate the necessity of the proxy estimation network by conducting an experiment that excludes the proxy estimation module and estimates the normal map with only face images as input in the normal estimation network. Secondly, we show the effectiveness of data augmentation for training the proxy estimation network, with another experiment that trains the proxy estimation network without the $5000$ synthesized images derived from data augmentation. The comparison results on test set for both experiments are shown in Tab.~\ref{table:ablation_study} and Fig.~\ref{fig:ablation_study}. We can see that excluding each component will cause a drop performance for both three image inputs and single image input.

\begin{table}[!t]
	\centering  
	\caption{Average angular errors (in degrees) on test set for ablation studies.}	
	\label{table:ablation_study}
	\centering
	\begin{tabular}{cccc}  
		\toprule   
		\# Input & w/o Proxy & w/o Augmentation & Ours \\  
		\midrule   
		1 & 14.843 & 12.342 & \textbf{9.875}  \\  
		3 & 9.499 & 8.694 & \textbf{6.154}  \\
		\bottomrule  
	\end{tabular}
\end{table}

\subsection{Comparisons}

\begin{figure}[!t]
	\centering
	\includegraphics[width=\columnwidth]{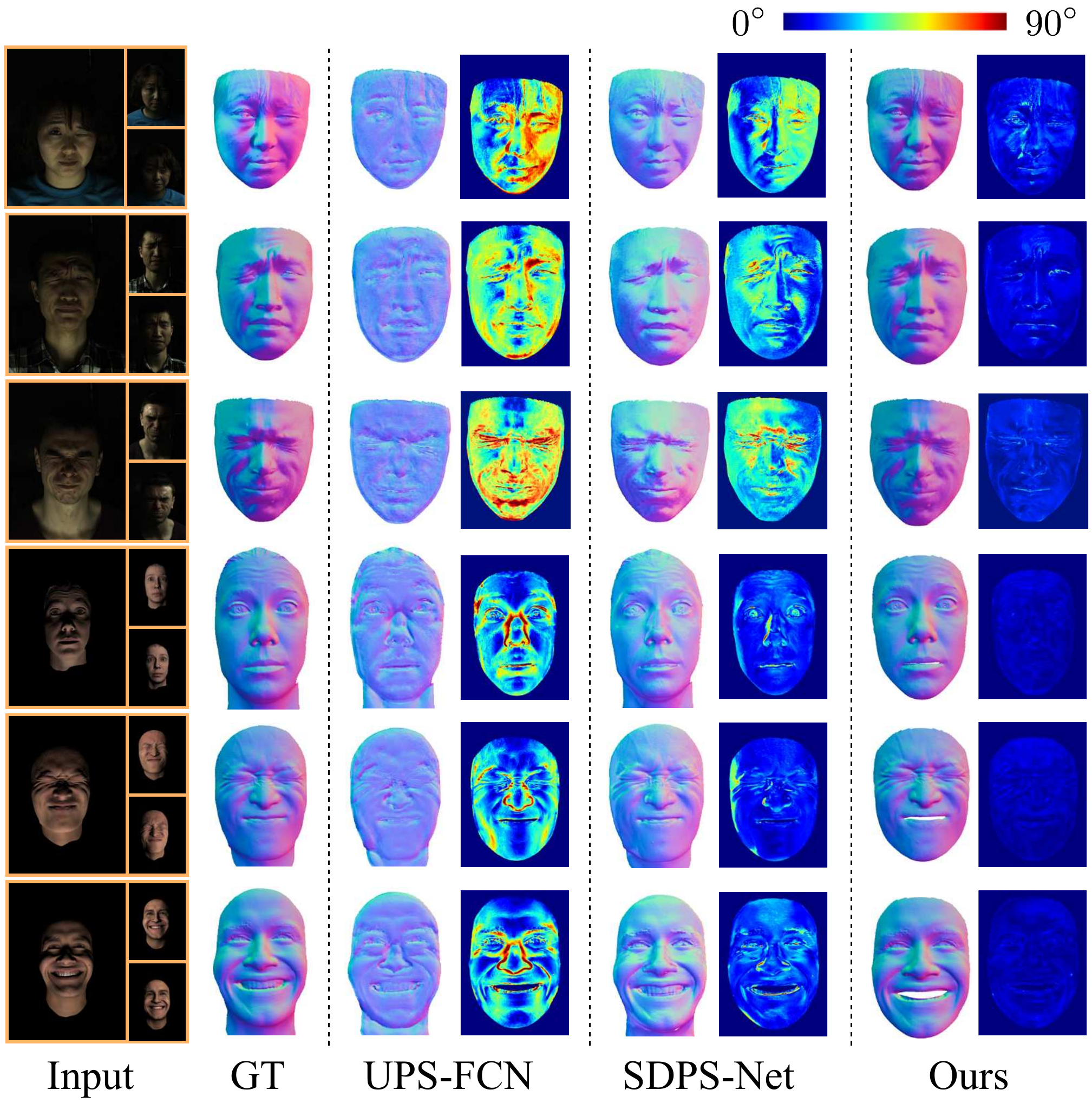}
	\caption{Estimated normal maps and their corresponding error maps for UPS-FCN~\cite{chen2018ps}, SDPS-Net~\cite{chen2019self} and our network.}
	\label{fig:PS_results}
\end{figure}

\paragraph{Comparison with deep learning-based photometric stereo.} We further compare our network with UPS-FCN~\cite{chen2018ps} and SDPS-Net~\cite{chen2019self} that solve the uncalibrated photometric stereo problem. Both methods estimate normals for general objects under different directional lights, whereas we focus on the human face under different point lighting conditions. We take three images with different uncalibrated lighting conditions from the test set as input and compare the accuracy of the output normal map according to the angle between the output and the ground truth normal map. We show the results in Table.~\ref{table:PS_results} and Fig.~\ref{fig:PS_results}. It can be observed from Table.~\ref{table:PS_results} that all methods perform better on the Light State test data, potentially due to noises in the real captured data. On the other hand, our method performs better than the other two methods both qualitatively and quantitatively, due to the near point lighting hypothesis and the face prior information.
\begin{table}[!t]
	\centering  
	\caption{Average angular errors (in degrees) on test sets.}	
	\label{table:PS_results}
	\centering
	\begin{tabular}{cccc}  
		\toprule   
		& UPS-FCN &SDPS-Net & Ours \\  
		\midrule   
		Real Set & 45.708 & 33.154 & \textbf{6.579}  \\  
		Light Stage~\cite{ma2007rapid} & 31.254 & 15.592 & \textbf{5.007}  \\
		\bottomrule  
	\end{tabular}
\end{table}

\begin{figure}[!t]
	\centering
	\includegraphics[width=1\columnwidth]{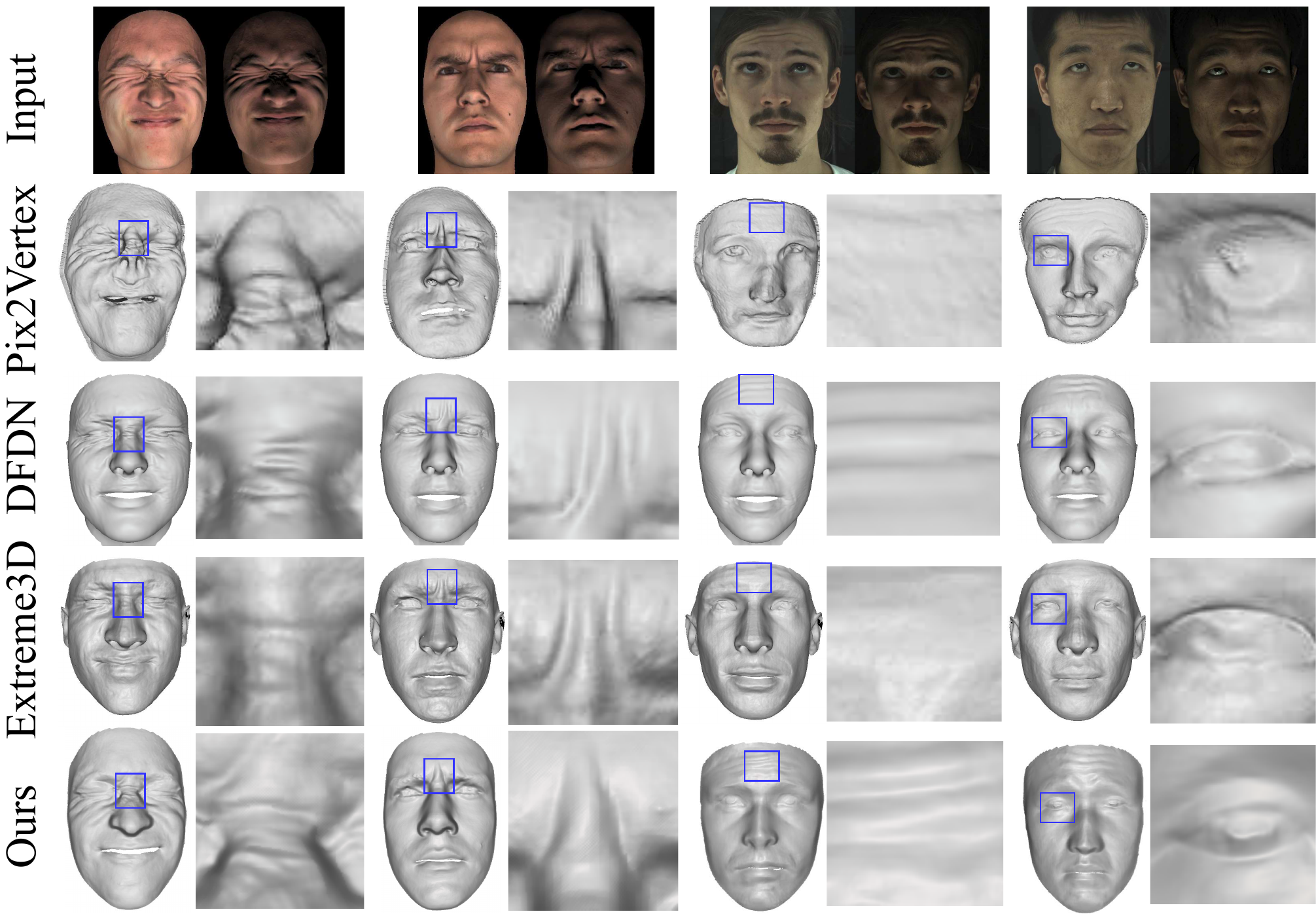}
	\caption{Qualitative comparison between Pix2vertex~\cite{sela2017unrestricted}, DFDN~\cite{chen2019photo}, Extreme3D~\cite{tran2018extreme} and our method. Other methods use the left image on the top row as input while ours uses the right image as input. Our method can reconstruct more accurate face models with fine details such as wrinkles and eyes.}
	\label{fig:single_quality_result}
\end{figure}

\begin{figure}[!t]
	\centering
	\includegraphics[width=1\columnwidth]{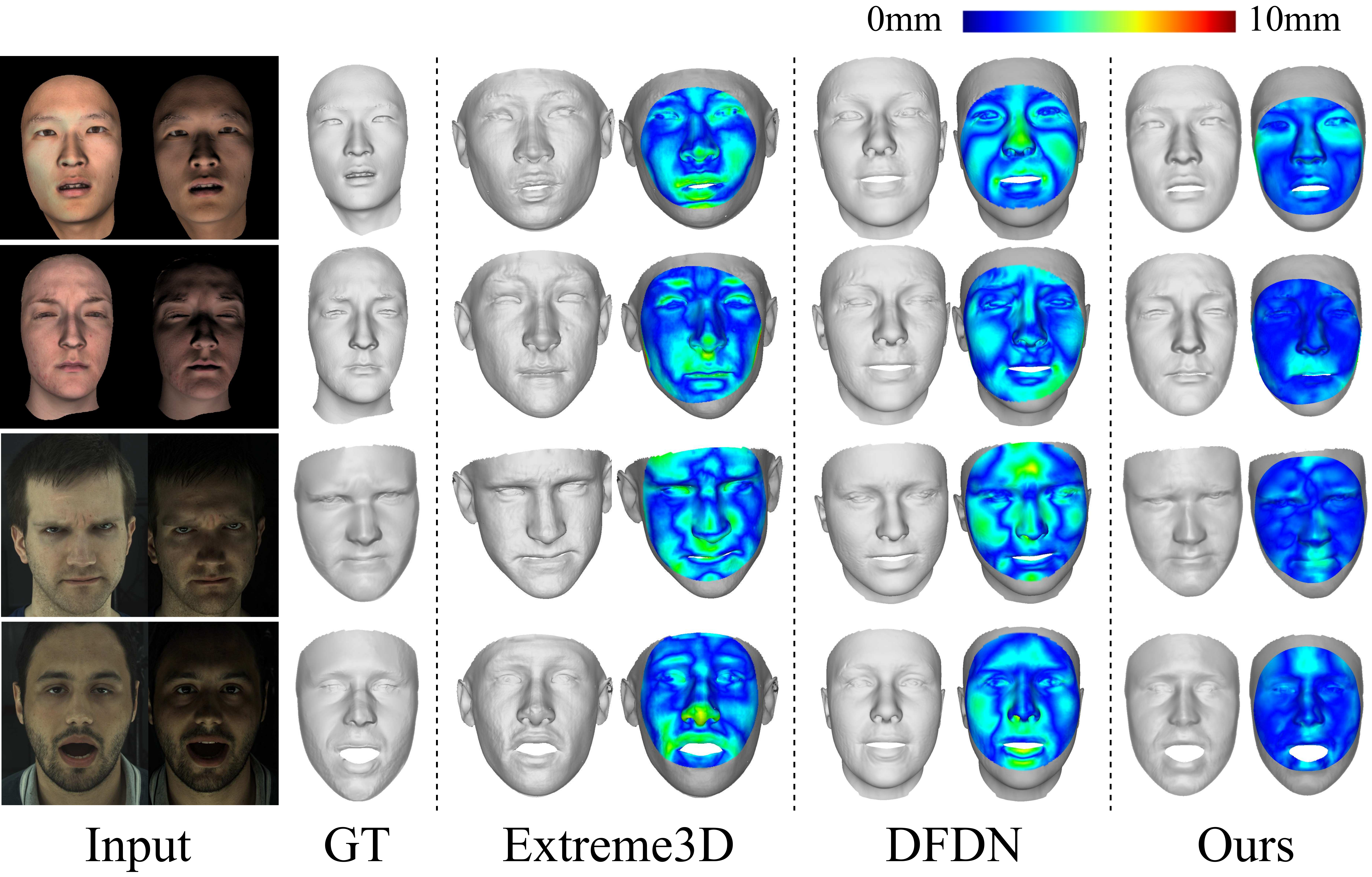}
	\caption{Reconstructed results and geometric error maps of Extreme3D~\cite{tran2018extreme}, DFDN~\cite{chen2019photo} and ours. Other methods use the left image in the first column as input while ours uses the right image as input.}
	\label{fig:geometric_error}
\end{figure}

\paragraph{Comparison with 3D face reconstruction from a single image.} In order to evaluate the quality of our reconstructed 3D face models, we compare our deep learning-based reconstruction method with some state-of-the-art detail-preserving reconstruction methods from a single image. Most existing methods focus on reconstruction from an ``in-the-wild'' image and simulate the environment lighting condition using the spherical harmonics (SH) basis functions, which performs poorly in simulating the near point lighting condition due to a large area of shadows. For a fair comparison, we take only one photometric stereo image as input to our network and one image captured in normal light as input to compared methods. The results shown in Fig.~\ref{fig:single_quality_result} demonstrate that our method can better recover facial details such as wrinkles and eyes. For quantitative evaluation, we compute a geometric error for each reconstructed model, by first applying a transformation with seven degrees of freedom (six for rigid transformation and one for scaling) to align it with the ground-truth model, and then computing its point-to-point distance to the ground-truth model. The average geometric errors of Extreme3D~\cite{tran2018extreme}, DFDN~\cite{chen2019photo} and our method on test set are $1.77$, $1.54$, $0.86$ respectively, with four examples shown in Fig.~\ref{fig:geometric_error}. It can be seen that our method significantly outperforms other methods due to our accurate simulation of the near point lighting condition.

\section{Conclusion}

We proposed a lightweight photometric stereo algorithm combining deep learning method and face shape prior to reconstruct 3D face models containing fine-scale details. Our two-stage neural network estimates a coarse face shape with structure and a normal map with details, followed by an optimization method to recover the final facial geometry. For the network training, we construct a real dataset across different races, genders and ages, and a data augmentation is applied to enrich the dataset. Extensive experiments demonstrated that our method outperforms state-of-the-art deep learning-based photometric stereo methods and 3D face reconstruction methods from a single image.

\noindent \textbf{Acknowledgement}
This work was supported by the National Natural Science Foundation of China (No. 61672481), and Youth Innovation Promotion Association CAS (No. 2018495).

{\small
\bibliographystyle{ieee_fullname}
\bibliography{egbib}
}

\section*{Appendix}

\subsection*{Test on ``In-The-Wild'' Images}

\begin{figure}[b]
	\centering
	\includegraphics[width=0.95\columnwidth]{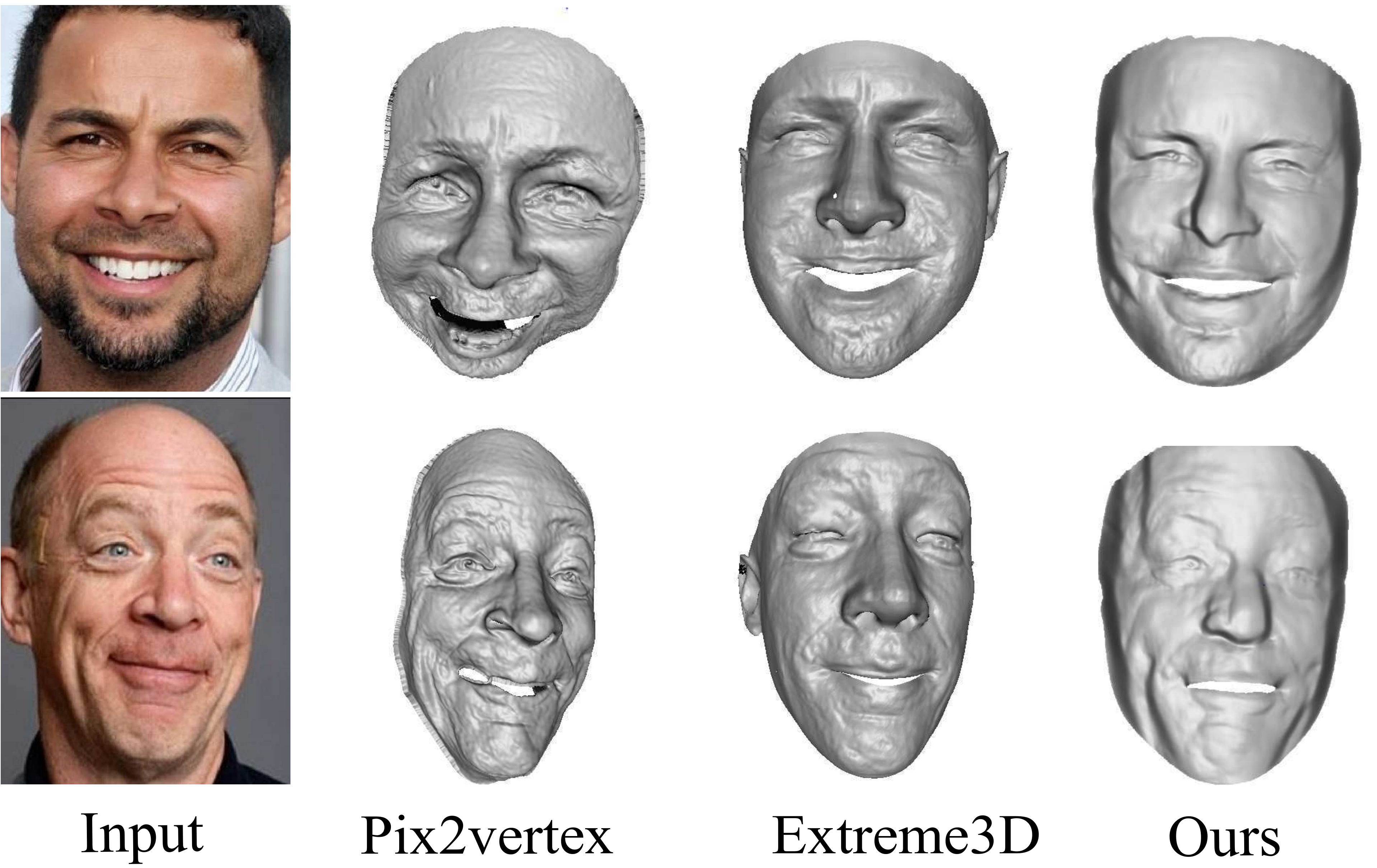}
	\caption{Qualitative comparison among Pix2vertex [36], Extreme3D [41] and our method on ``in-the-wild'' images.}
	\label{fig:supplement_wild}
\end{figure}

In this paper, we focus on high-quality 3D face reconstruction from face images captured with near-field point light source. Although our proposed method is not intended for ``in-the-wild'' images, we also test our method on such images and compare with Pix2vertex [36] and Extreme3D [41]. Fig.~\ref{fig:supplement_wild} demonstrates that our method achieves reasonable results on such images. It is possibly due to our proxy estimation network that utilizes facial shape priors to generate a 3D face model.

\subsection*{Results with Different Inputs}

\begin{figure*}[b]
	\centering
	\includegraphics[width=0.92\textwidth]{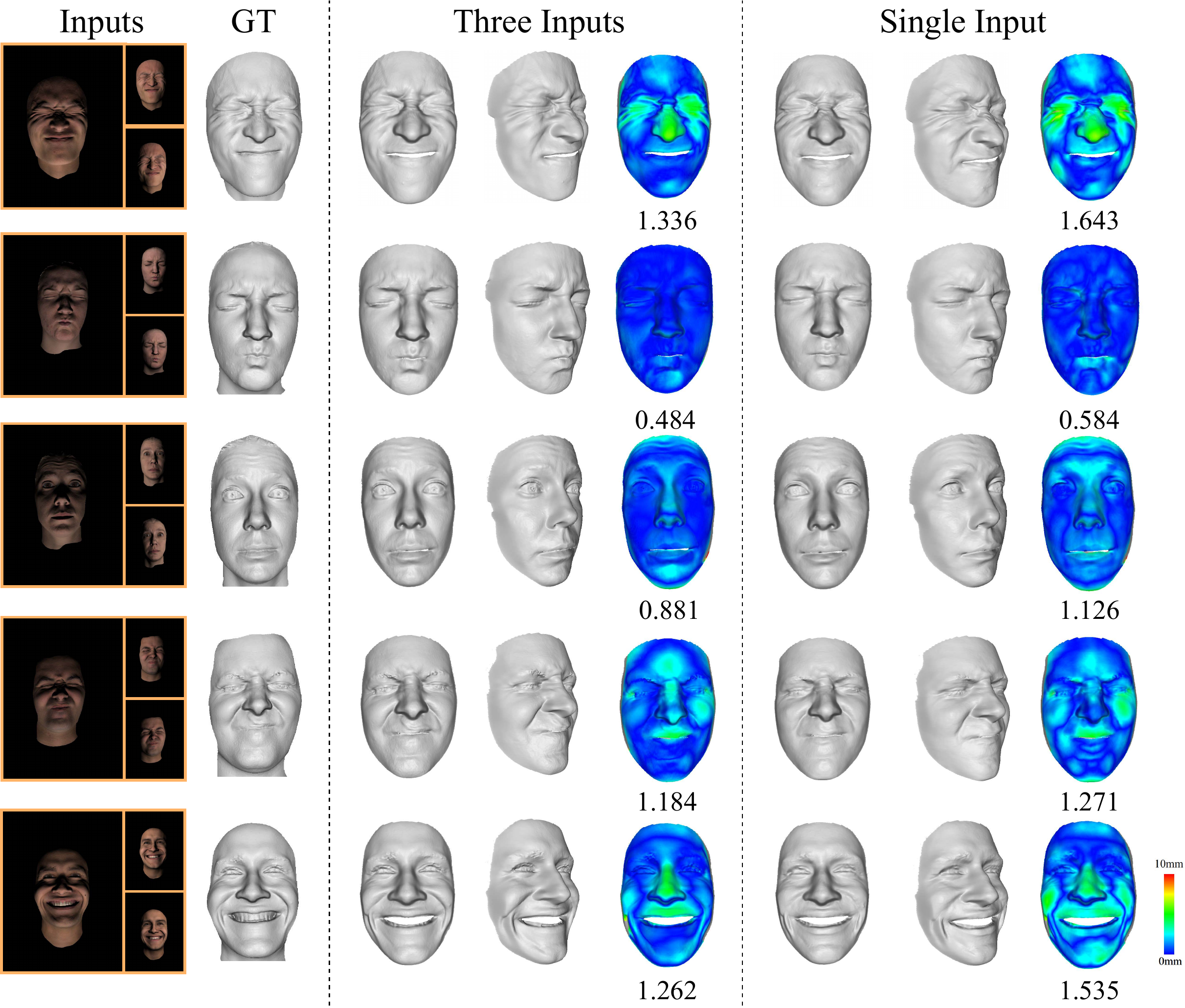}
	\caption{Reconstruction results on the Light Stage. Given the images as shown in the left, we show the reconstruction result under frontal view and side view. The corresponding error map and the geometric error of each reconstruction result are also given. The leftmost image is the single image input.}
	\label{fig:supplement_light}
\end{figure*}

\begin{figure*}[b]
	\centering
	\includegraphics[width=1\textwidth]{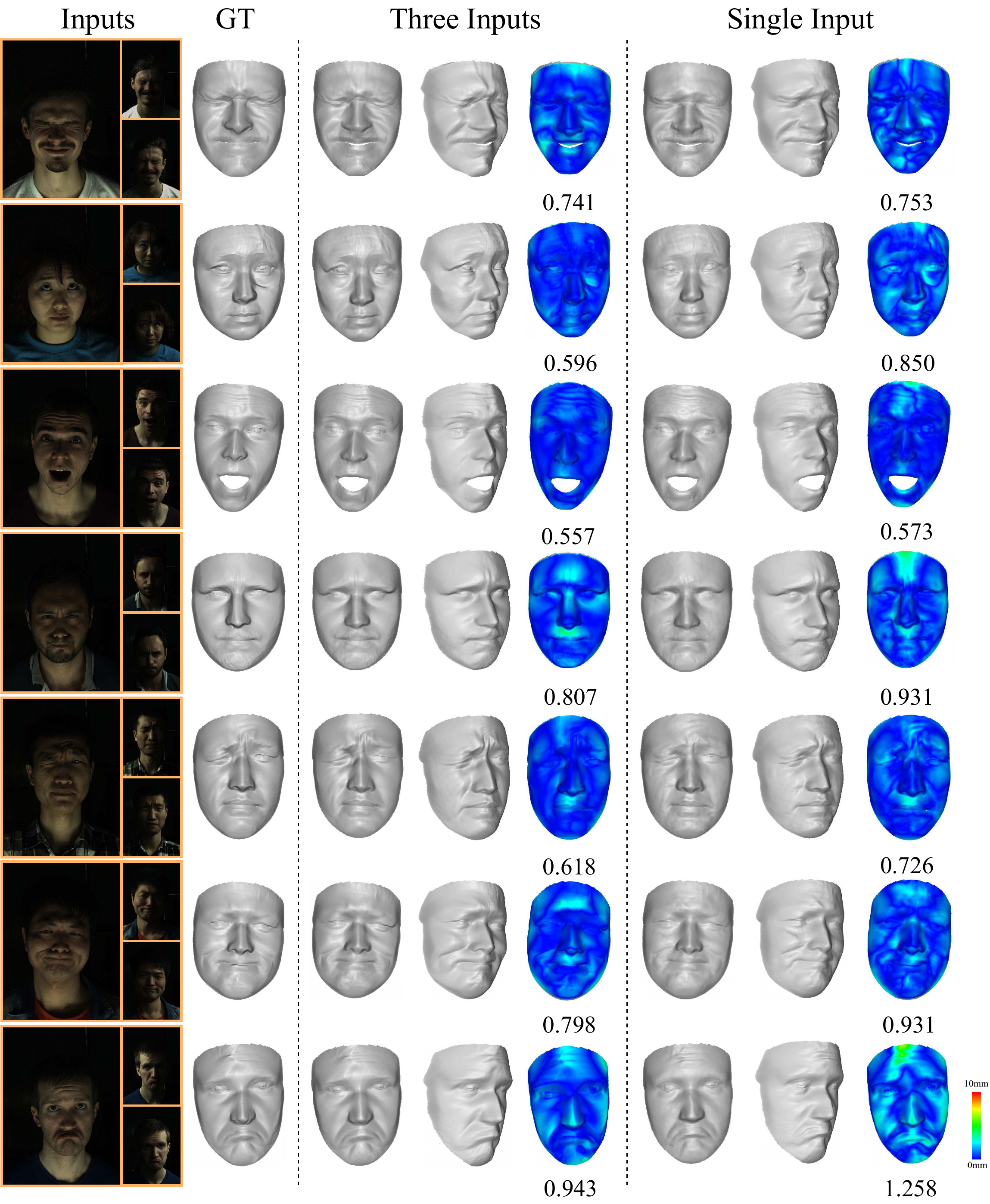}
	\caption{Reconstruction results on the real test set. Given the images as shown in the left, we show the reconstruction result under frontal view and side view. The corresponding error map and the geometric error of each reconstruction result are also given. The leftmost image is the single image input.}
	\label{fig:supplement_real}
\end{figure*}

\textbf{Three Images Input vs. Single Image Input.} Our method can reconstruct face shapes from arbitrary number of face image inputs. In Fig.~\ref{fig:supplement_light} and Fig.~\ref{fig:supplement_real}, we show reconstruction results of all the 12 subjects in our set (seven from the real set and five from the Light Stage) with three input images and a single image respectively. The results show that our method can recover fine facial details from both settings of inputs, and better accurate reconstruction results can be achieved from three input images due to the richer information provided by more inputs.


\begin{figure*}
	\centering
	\includegraphics[width=1.0\textwidth]{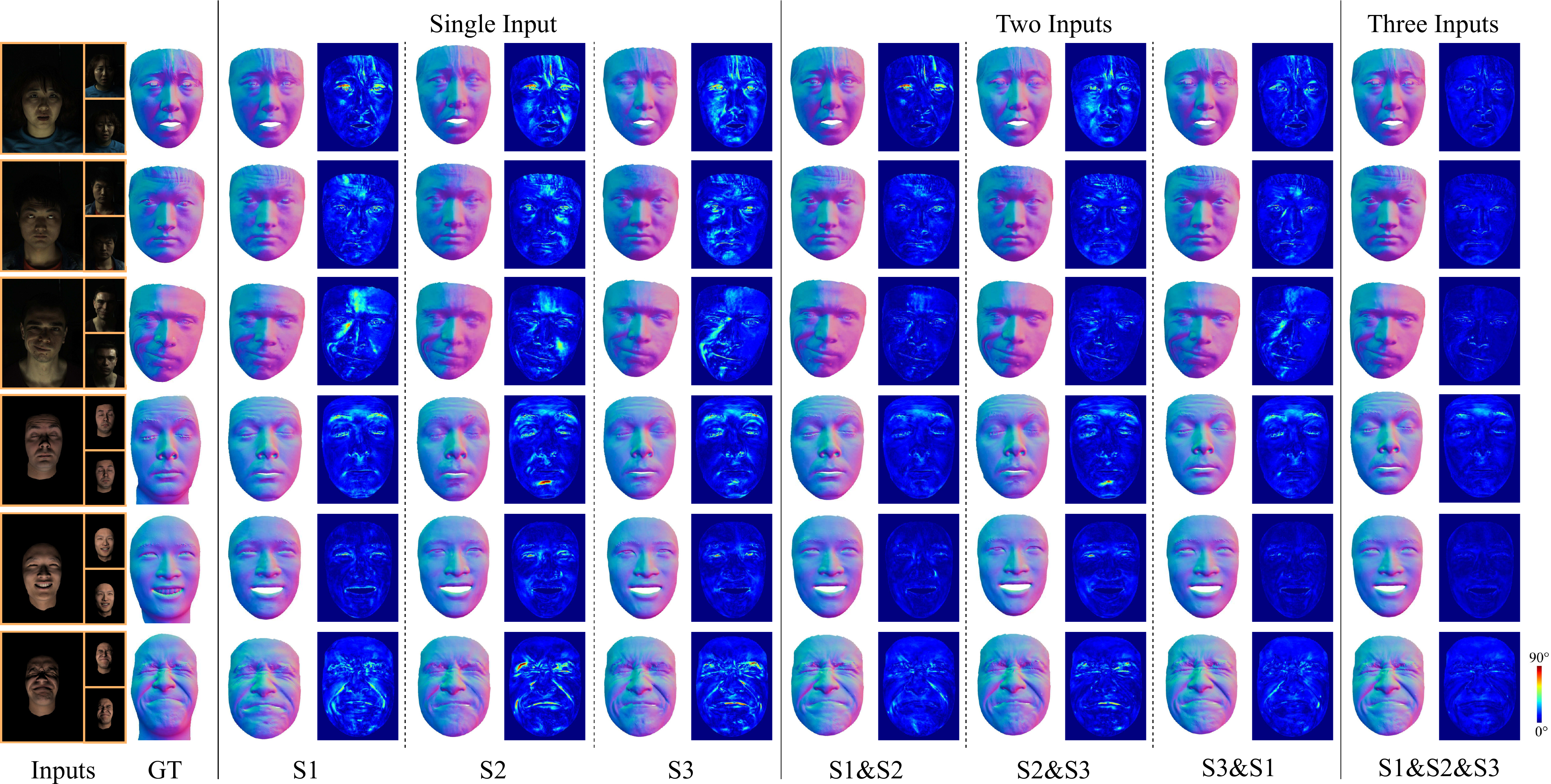}
	\caption{Estimated normal maps. For each model we show normal maps and their corresponding error maps with different kinds of inputs. \emph{S1, S2, S3} represent the leftmost, the upper-right corner and the lower-right corner image respectively.}
	\label{fig:supplement_nettest}
\end{figure*}

\textbf{Arbitrary Images Input.} In the paper we quantitatively show the results with arbitrary inputs of our proposed method. More results on our test set with arbitrary inputs are shown in Fig.~\ref{fig:supplement_nettest}. It can be observed that more input images lead to more accurate results.

\end{document}